\documentclass[runningheads]{llncs}

\usepackage{eccvabbrv}

\usepackage{graphicx}
\usepackage{booktabs}
\usepackage{multirow}

\usepackage[accsupp]{axessibility}  

\usepackage{hyperref}
\usepackage{cleveref}

\usepackage{orcidlink}

\begin{document}

\title{GEAR-Seg: A Grounded Explainable Agent for Reasoning Segmentation and Data Engine} 

\titlerunning{GEAR-Seg: A Grounded Explainable Agent for Reasoning Segmentation}



\author{Yanan Wang\inst{1}\orcidlink{0009-0003-3303-7440} \and
Wen Li\inst{1}\orcidlink{0009-0005-6453-2924} \and
Yibin Ying\inst{1,2}\orcidlink{0000-0002-3392-9380}\and
Zhenghao Fei\inst{1,2}\orcidlink{0000-0001-9132-354X}\thanks{Corresponding author.}}

\authorrunning{Y.~Wang et al.}

\institute{College of Biosystems Engineering and Food Science, Zhejiang University, Hangzhou 310058, China \and
ZJU-Hangzhou Global Scientific and Technological Innovation Center, Zhejiang University, Hangzhou 311215, China
\email{\{mmwang, zfei\}@zju.edu.cn}}

\maketitle



\begin{figure*}[t]
  \centering
  \includegraphics[width=1\textwidth]{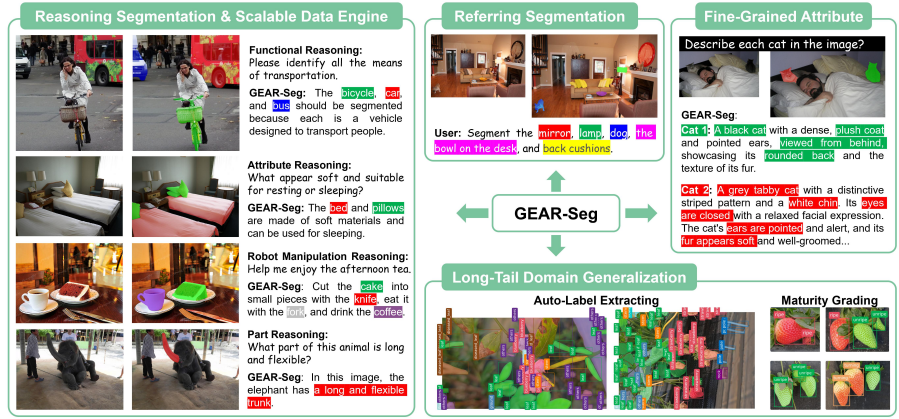}
  \caption{Overview of GEAR-Seg's multifaceted capabilities. Serving as both a zero-shot inference agent and a scalable data engine, it explicitly translates pixels into text to seamlessly support complex reasoning segmentation, dense referring segmentation, and fine-grained attribute grounding in long-tail domains.}
  \label{fig:overview}
\end{figure*}

\begin{abstract}
  Reasoning segmentation requires localizing targets based on complex, implicit queries. Current end-to-end models typically entangle perception and deduction into an opaque black box, severely limiting interpretability and scalability. To address this, we propose GEAR-Seg (\textbf{G}rounded \textbf{E}xplainable \textbf{A}gent for \textbf{R}easoning \textbf{S}egmentation), an explicitly decoupled agent that shifts the paradigm by translating visual pixels into dense, attribute-rich text. By decoupling class-agnostic segmentation, semantic description, and Large Language Model (LLM) deduction, GEAR-Seg transforms implicit reasoning into an explicit, trackable logic chain. As a zero-shot inference framework, it achieves highly competitive performance across diverse reasoning and fine-grained referring segmentation benchmarks. Furthermore, GEAR-Seg inherently functions as a highly scalable data engine. Utilizing this engine, we construct GEAR-131K, a massive benchmark (over 38k images, 656k QA-mask pairs) introducing a multifaceted taxonomy tailored for complex real-world manipulation-oriented reasoning. Finally, distillation experiments demonstrate that lightweight models supervised exclusively by our automated pipeline closely match the upper-bound performance of costly human-annotated baselines.
\end{abstract}

\section{Introduction}
\label{sec:intro}
Visual segmentation has witnessed a remarkable evolution, transitioning from standard closed-set pixel classification \cite{unet2015,sert2021,minderer2022simple} to open-vocabulary recognition \cite{groundedsam_2024,yolo-world_2024}, and most recently, to reasoning segmentation \cite{lavt2023,lisa2024}. Unlike traditional referring expression comprehension \cite{rec2016,mattnet2018} that relies on explicit noun phrases, reasoning segmentation requires models to accurately localize target regions based on complex, implicit textual queries. This task demands a deep fusion of visual perception and high-level cognitive deduction, making it a challenging benchmark for evaluating visual grounding.

Despite rapid progress, current state-of-the-art (SOTA) architectures typically formulate reasoning segmentation as an end-to-end framework. By directly injecting visual features into LLMs, these methods \cite{lisa2024,llmseg2024,ICLR2025_MMR} implicitly entangle perception and reasoning into an opaque black box. This architectural choice inherently restricts interpretability, limits zero-shot generalization, and creates a profound reliance on expensive human annotations. Consequently, existing benchmarks \cite{lisa2024,llmseg2024} remain severely limited in diversity, often confining queries to isolated instances or homogeneous categories (e.g., targeting "the single hat" or "all hats" in the image).  
Most critically, they lack a systematic taxonomy covering diverse reasoning scenarios beyond commonsense grounding. For instance, executing a cross-category instruction like \textit{"divide the dessert into small pieces"} demands profound semantic comprehension to deduce the implicit targets (e.g., identifying the "dessert" as the cake and inferring the required tool as a knife), coupled with the precise spatial localization of these interacting entities---a complex deductive capability largely absent in current datasets.

To address these critical bottlenecks, we propose GEAR-Seg (\textbf{G}rounded \textbf{E}xplainable \textbf{A}gent for \textbf{R}easoning \textbf{S}egmentation), a novel framework that shifts the paradigm from implicit entanglement to explicit modular decoupling. As shown in \cref{fig:overview}, at its core, GEAR-Seg pioneers a powerful pixel-to-text translation strategy. It decomposes the task into three transparent stages: universal region proposal via the Segment Anything 2 (SAM 2) \cite{SAM2}, dense semantic description of every mask using the specialized Describe Anything Model (DAM) \cite{dam2025}, and multi-step deduction driven by a pure LLM. By explicitly turning pixels into rich, human-readable text, and utilizing pretrained knowledge in the LLM, this decoupled architecture bypasses the information bottlenecks of previous methods. It equips GEAR-Seg with dual operational modes: addressing complex reasoning segmentation via abstract deduction, and serving as a formidable dense referring segmentation engine via prompt-constrained grounding. Leveraging these unconstrained visual contexts and detailed physical attributes, GEAR-Seg achieves competitive zero-shot inference performance. Notably, it demonstrates strong generalization in long-tail applications, effectively handling fine-grained real-world tasks such as agricultural maturity grading without requiring manual annotations.

Beyond serving as a robust inference framework, the explicitly decoupled and transparent nature of GEAR-Seg uniquely positions it as a highly scalable data engine. To address the severe data scarcity for downstream deployment, we leverage GEAR-Seg to automatically construct GEAR-131K, a massive reasoning segmentation dataset comprising over 38k images and 656k diverse QA-mask pairs. Breaking away from restrictive ``commonsense-only'' norms, our dataset introduces a comprehensive taxonomy tailored for complex interactions: Commonsense, Functional, Manipulation-related, Part-based, and Attribute-based reasoning. To validate the high fidelity of our automated supervision, we train end-to-end networks \cite{lisa2024} exclusively on our generated data. Extensive experiments demonstrate a profound data-centric advantage: models supervised solely by our low-cost GEAR-Seg pipeline easily surpass the performance of baselines trained on expensive human-annotated datasets.

In summary, our contributions are threefold:
\begin{enumerate}
\item \textbf{A Modular Agent with Pixel-to-Text Translation:} We propose GEAR-Seg, an explicitly decoupled agent that shifts the field from implicit end-to-end entanglement to transparent, modular reasoning. By systematically replacing sparse category prompts with dense, mask-level attribute descriptions via a \textit{pixel-to-text} paradigm, GEAR-Seg achieves competitive zero-shot performance in reasoning segmentation and dense referring segmentation.
\item \textbf{A Scalable Data Engine and Comprehensive Benchmark:} Leveraging GEAR-Seg as an automated generation pipeline, we construct GEAR-131K, a massive, high-fidelity reasoning segmentation dataset (over 38k images, 656k QA-mask pairs). Breaking away from traditional single-target constraints, our dataset introduces a multi-dimensional taxonomy strictly tailored for real-world manipulation-oriented reasoning and functional affordances. 
\item \textbf{Effective Knowledge Distillation Across Model Scales:} We propose a low-cost, data-centric distillation paradigm. By training end-to-end networks exclusively on our automatically generated dataset, we effectively distill the cognitive capabilities of the GEAR-Seg agent into downstream architectures. Extensive experiments demonstrate that student models---ranging from large VLMs (e.g., LISA \cite{lisa2024}) to lightweight models (e.g., YOLOv8 \cite{YOLO-v8})---supervised solely by our generated data achieve highly competitive performance, closely matching the rigorous upper-bound metrics of their counterparts trained on costly human-annotated datasets.

\end{enumerate}

\section{Related Work}
\subsection{Evolution from Referring to Reasoning Segmentation}
Visual segmentation has evolved from predefined category recognition to open-ended, language-driven perception \cite{SAM_2023}. While referring segmentation methods (e.g., VLT \cite{VLT}, CRIS \cite{CRIS}) and open-vocabulary models (e.g., Grounded SAM \cite{groundedsam_2024}, YOLO-World \cite{yolo-world_2024}) excel at localizing targets specified by explicit nouns, they fundamentally struggle with implicit queries requiring the deduction of human intents or physical affordances. To bridge this gap, recent works have introduced reasoning segmentation. Pioneering models like LISA \cite{lisa2024}, LLM-Seg \cite{llmseg2024}, and PixelLM \cite{pixellm2024} address this by injecting visual tokens into LLMs to perform deduction and segmentation end-to-end. However, these methods implicitly entangle perception and reasoning within an opaque black box. This architectural limitation not only restricts interpretability but also degrades performance in long-tail domains that demand fine-grained attribute differentiation \cite{longtail2020}. Consequently, there is an urgent need for paradigms that explicitly decouple visual perception from cognitive deduction.

\subsection{Benchmarks for Reasoning Segmentation}
Due to the emerging nature of reasoning segmentation, high-quality and comprehensive datasets remain relatively scarce \cite{lisa2024,llmseg2024,ICLR2025_MMR}. Although recent benchmarks have expanded task boundaries from single-instance targets to multi-category scenarios, they predominantly lack designs tailored for embodied AI and complex robotic manipulation \cite{pixellm2024}. Early benchmarks, such as ReasonSeg \cite{lisa2024}, rely heavily on costly human annotations, severely limiting their scale. To alleviate this, recent works explore automated data generation pipelines; for instance, LLM-Seg \cite{llmseg2024} employs GPT for prompt generation, and PixelLM \cite{pixellm2024} utilizes LLaMA for brief target descriptions. Nevertheless, these methods are bottlenecked by their reliance on predefined vocabularies or existing dataset annotations, failing to provide the LLM with an unbiased, holistic scene understanding. Furthermore, existing benchmarks predominantly provide simple binary (Query, Mask) pairs or rudimentary category lists, completely lacking the structured, logical reasoning explanation chains crucial for training interpretable models.

\subsection{Agentic AI and Knowledge Distillation}
The rapid advancement of LLMs has catalyzed the emergence of Agentic AI \cite{AgenticAI_survey}, characterized by its ability to explicitly decompose complex, open-ended problems into manageable sub-tasks \cite{agents_vs_agentic2025}. Visual agents leverage LLMs as central cognitive brains to orchestrate specialized tools via API calls \cite{SAM4MLLM,cogagent2023}, achieving impressive zero-shot capabilities. However, deploying such heavy, multi-step agents in real-time or resource-constrained robotic systems is highly impractical \cite{kozlov2021neural}. To overcome this, data-centric knowledge distillation \cite{sachdeva2022infinite} has emerged as an effective paradigm. Instead of directly distilling model weights, this approach utilizes a powerful agent as a teacher to automatically generate high-fidelity training data, which subsequently supervises lightweight, end-to-end student models \cite{sdmd2026}. Building upon these insights, we introduce GEAR-Seg. Operating initially as a transparent, decoupled agent, it overcomes previous architectural limitations to achieve state-of-the-art zero-shot reasoning. Crucially, it doubles as a highly scalable data engine to automatically construct a massive benchmark---complete with dense explanation chains---thereby seamlessly distilling its advanced cognitive capabilities into various downstream architectures.

\section{Method}

\subsection{The GEAR-Seg Framework}

Given an input image $x$ and an implicit reasoning query $Q$, our objective is to predict a precise segmentation mask $M$ that satisfies the complex logical intent embedded within $Q$. To achieve this, we propose the GEAR-Seg framework, as illustrated in Fig.~\ref{fig:method}. Unlike end-to-end black-box approaches, GEAR-Seg operates across three explicitly decoupled modules:

\begin{figure*}[t]
  \centering
  \includegraphics[width=1\textwidth]{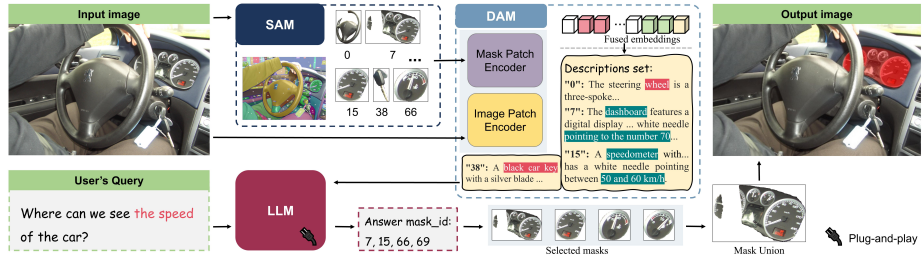}
  \caption{Overview of the GEAR-Seg framework. The agent explicitly decouples the reasoning segmentation task into class-agnostic perception (SAM 2), dense semantic description (DAM), and logic-driven abstraction (LLM), serving as both a zero-shot inference engine and a scalable data generator.}
  \label{fig:method}
\end{figure*}

\textbf{1) Class-Agnostic Segmentation.} We employ SAM 2 \cite{SAM2} in Everything Mode. A Mask Non-Maximum Suppression \cite{sdmd2026} yields valid instance masks $\mathcal{M} = \{m_1, m_2, \dots, m_N\}$, ensuring no salient object or subtle background attribute is overlooked.

\textbf{2) Dense Semantic Description.} This is the core pixel-to-text paradigm shift. Using DAM \cite{dam2025}, the global context $z^G$ and localized features $z^R$ are extracted. DAM fuses these to generate a context-aware description $d_i$ for each mask, producing a structured sequence $\mathcal{D} = \{d_1, d_2, \dots, d_N\}$. This fine-grained text is the interpretable bedrock for reasoning.

\textbf{3) Logic-Driven Reasoning.} Bypassing entangled VLMs, we deploy a plug-and-play LLM. Given $Q$ and $\mathcal{D}$, the LLM explicitly evaluates semantic entailment to output valid mask indices $I = \mathcal{F}_{\text{LLM}}(Q, \mathcal{D})$. The final output is $\widehat{M} = \bigcup_{i \in I} m_i$. And this decoupled architecture endows GEAR-Seg with two synergistic modes: \textbf{Reasoning Segmentation Mode}, where the LLM autonomously discovers targets from implicit queries via abstract deduction; and \textbf{Referring Segmentation Mode}, where fine-grained physical attributes in $\mathcal{D}$ are exploited to ground explicit instructions, significantly outperforming open-vocabulary SOTAs in long-tail tasks.

\subsection{Data-Centric Knowledge Distillation}
To facilitate edge deployment, GEAR-Seg functions as a scalable data engine via a pure data distillation paradigm. Crucially, the student and teacher models remain architecturally independent. Given unannotated images $\mathcal{X} = \{x_1, \dots, x_K\}$, GEAR-Seg generates diverse queries $Q_g^k$ alongside pseudo-labels $Y_p^k$ (comprising both explicit logic chains and segmentation masks):
\begin{equation}
Y_p^k = \text{GEAR-Seg}(x_k, Q_g^k)
\label{eq:pseudo_label}
\end{equation}
These labels supervise a lightweight student $f_\theta$ entirely through its \textit{native} task-specific objective $\mathcal{L}$:
\begin{equation}
\theta^* = \arg \min_{\theta} \sum_{k=1}^{K} \mathcal{L}\left(f_{\theta}(x_k, Q_g^k), Y_p^k\right)
\label{eq:student_opt}
\end{equation}
Crucially, the loss $\mathcal{L}$ refers entirely to the student's \textit{native} objective. For instance, in student Vision-Language Models (e.g., LISA), this native loss inherently comprises both text and mask objectives ($\mathcal{L} = \mathcal{L}_{txt} + \mathcal{L}_{mask}$). This flexible design allows task-specific lightweight models to genuinely inherit GEAR-Seg's deductive strengths while achieving the rapid inference latency required for robotics.

\section{The GEAR-131K Benchmark}

Beyond zero-shot inference, GEAR-Seg inherently functions as a scalable data engine. To address the scarcity of complex interaction scenarios in existing benchmarks, we construct GEAR-131K, a massive dataset generated via our automated pipeline.

\subsection{Data Engine Workflow}

\begin{figure}[t]
  \centering
  \includegraphics[width=\textwidth]{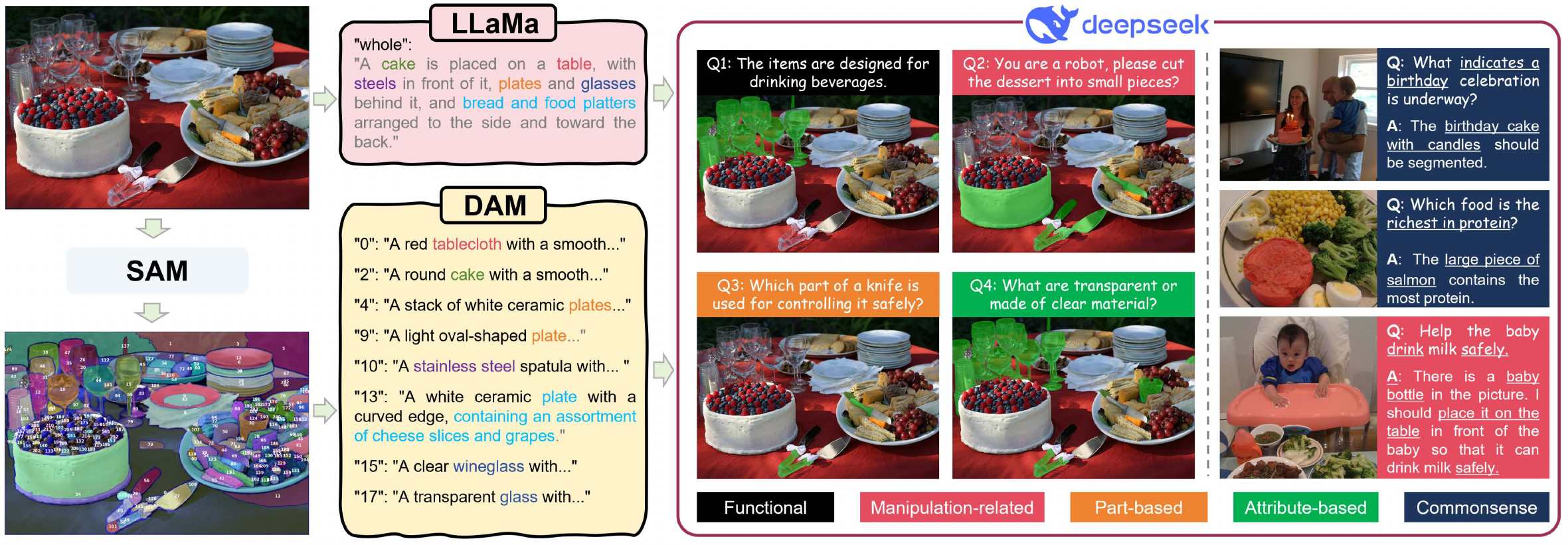}
  \caption{Overview of the GEAR-Seg data generation pipeline and operational modes.}
  \label{fig:dataset_overview}
\end{figure}

As illustrated in Fig.~\ref{fig:dataset_overview}, the creation of GEAR-131K leverages the GEAR-Seg framework as an autonomous, closed-loop data engine. The generation pipeline operates by fusing multi-granularity visual semantics. First, macro-level scene contexts and global relationships are extracted using a lightweight VLM (LLaMA 3.2~\cite{grattafiori2024llama3herdmodels}). Concurrently, at the micro-level, SAM~2 and the Dense Attribute Module (DAM) extract unconstrained, fine-grained semantic regions to capture intricate object details and attributes. Finally, a central LLM acts as the reasoning core, synthesizing these comprehensive modalities to autonomously generate a diverse set of annotations. For each image, the engine outputs a challenging base query, an explicit step-by-step logic chain, and the corresponding precise mask indices, thereby establishing a high-quality benchmark for reasoning segmentation.

\subsection{Taxonomy of Reasoning Scenarios}
To bridge the critical gap between standard academic evaluation and complex real-world physical interactions, GEAR-131K pioneers the explicit categorization of multifaceted reasoning scenarios. We systematically break traditional single-target limitations by defining five distinct reasoning categories:
\begin{itemize}
    \item \textbf{Commonsense Reasoning:} Evaluates deductive capabilities based on contextual visual cues (e.g., identifying the birthday cake from \textit{``What suggests a birthday celebration here?''}).
    \item \textbf{Functional Reasoning:} Targets object groups based on utility or affordances, breaking single-category limitations (e.g., retrieving cars, bicycles, and buses simultaneously given \textit{``Identify all means of transportation''}).
    \item \textbf{Manipulation-related:} Designed for complex interaction tasks requiring multi-target coordination across distinct semantic categories (e.g., returning both the cake and knife for \textit{``Divide the dessert into several pieces''}).
    \item \textbf{Part-based Reasoning:} Breaks the holistic object-level boundary by localizing fine-grained, sub-instance regions, demanding deeper structural understanding (e.g., isolating \textit{``the safe-to-hold part of the knife''}).
    \item \textbf{Attribute-based Reasoning:} Uniquely enabled by the rich dense descriptions of GEAR-Seg, this category evaluates the fine-grained perception of physical properties (e.g., material, shape, state). This is crucial for embodied agents where physical attributes dictate safety and affordances (e.g., distinguishing a \textit{``plastic toy knife''} from a \textit{``sharp metallic blade''}).
\end{itemize}

\subsection{Dataset Statistics and Integrity Protocol}

\begin{figure}[t]
    \centering
    \includegraphics[width=0.9\linewidth]{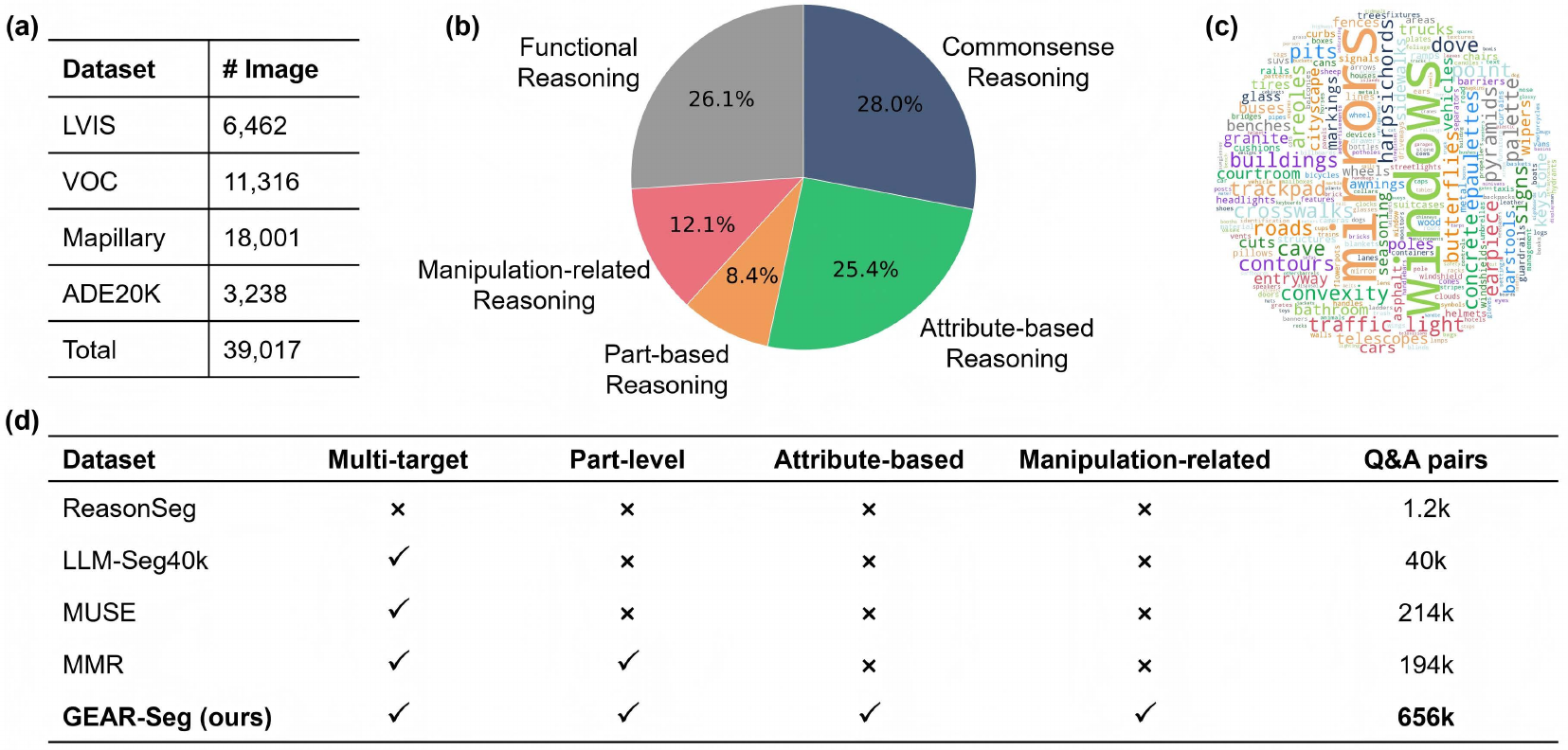}
    \caption{Detailed statistics of the GEAR-131K benchmark. (a) Image distribution across source datasets. (b) Proportion of the five specialized reasoning categories. (c) Word cloud illustrating the semantic diversity of the targeted entities. (d) Comprehensive feature comparison against existing reasoning segmentation datasets.}
    \label{fig:dataset}
\end{figure}

\textbf{Scale and Distribution.} As detailed in Fig. \ref{fig:dataset}(a), GEAR-131K comprises 39,017 high-resolution images meticulously curated from LVIS~\cite{Gupta2019LVIS}, VOC~\cite{VOC-_pascal_2010}, Mapillary~\cite{Neuhold_2017_ICCV}, and ADE20K~\cite{ADE20k}. Our automated engine initially generated 162k raw proposals. To guarantee high-fidelity supervision, we enforced rigorous automated filtering: discarding masks occupying $<1/600$ of the image area and removing pairs with LLM confidence scores $<6/10$. This refined the set to 131k base pairs (averaging 3.36 scenarios per image). Following a 5-fold linguistic expansion, the final dataset scales to over 656k QA-mask pairs, featuring a high-density average query length of 14.3 words. Furthermore, \cref{fig:dataset}(b) illustrates a balanced distribution across the five specialized reasoning scenarios, effectively mitigating the bias towards simplistic queries.

\textbf{Semantic Diversity and Superiority.} GEAR-131K encompasses 3,014 distinct target entities, as visualized in \cref{fig:dataset}(c), covering a vast spectrum of open-world categories, fine-grained parts, and unapparent physical attributes. Crucially, \cref{fig:dataset}(d) contextualizes our architectural advantages against existing benchmarks (ReasonSeg~\cite{lisa2024}, LLM-Seg40k~\cite{llmseg2024}, MUSE~\cite{pixellm2024}, MMR~\cite{ICLR2025_MMR}). GEAR-131K establishes a new comprehensive standard as the only benchmark simultaneously facilitating multi-target grounding, part-level reasoning, attribute-based differentiation, and complex manipulation-reltaed at scale.

\textbf{Data Splits and Annotation Protocol.} To prevent data leakage, we partition the dataset strictly at the image level into training (35.4k), validation (2k), and testing (1.8k) splits. Guaranteeing a rigorous, unbiased evaluation, the 1.8k-image test split underwent strict manual verification. Three independent experts reviewed the data following unified annotation guidelines, achieving a $>95\%$ inter-annotator agreement. Illogical queries were actively filtered, and inaccurate masks were corrected via Labelme, with edge-cases resolved through consensus. This ensures the test set serves as a highly reliable gold standard for complex embodied perception.

\section{Experiments}

\subsection{Zero-Shot Inference Performance of GEAR-Seg}
To systematically validate our framework, we evaluate its zero-shot inference capabilities across two synergistic operational modes. First, we assess the Reasoning Segmentation Mode using complex, implicit commonsense queries. Second, we evaluate the Referring Segmentation Mode through explicit, fine-grained attribute grounding in specialized, long-tail domains.

\subsubsection{Performance on Reasoning Segmentation Benchmarks.}
\label{subsubsec:reasoning}
We evaluate the agent's complex deductive capabilities across two representative public benchmarks.

\textbf{Evaluation Setup.} We utilize ReasonSeg \cite{lisa2024} and LLM-Seg40k \cite{llmseg2024} as evaluation datasets. We select SOTA reasoning models LISA \cite{lisa2024} and LLM-Seg \cite{llmseg2024}, and the referring method GRES \cite{gres2023} as baselines. Following standard protocols, we report gIoU \cite{rezatofighi2019giou} and cIoU \cite{zheng2020diou}. Additionally, due to the significant variance in image sizes within the ReasonSeg dataset, we report the normalized cIoU (ncIoU) \cite{zheng2020diou} to ensure a balanced evaluation.

\textbf{Results and Analysis.} As summarized in \cref{tab:reasonseg-tab}, GEAR-Seg demonstrates highly competitive zero-shot performance. On ReasonSeg, our zero-shot approach achieves a gIoU of 57.5, outperforming the fully fine-tuned LISA-13B (56.2). While GEAR-Seg yields a lower standard cIoU, it establishes superior performance on the size-balanced ncIoU metric (54.0 vs. 52.8). This divergence is an expected byproduct of our decoupled architecture: SAM 2 excels at precise sub-instance localization but occasionally over-segments massive targets in high-resolution images, disproportionately degrading unnormalized cIoU. By neutralizing this scale bias, the superior ncIoU reliably reflects GEAR-Seg's precise instance-level capabilities. Furthermore, on LLM-Seg40k, GEAR-Seg achieves a gIoU of 52.2, establishing a substantial margin over the fine-tuned LLM-Seg-7B (45.5). Some qualitative results are shown in \cref{fig:visual-results}(a).

\begin{table*}[t]
\centering
\caption{Quantitative comparison on the ReasonSeg and LLM-Seg40k validation splits. ``Use LLM'' indicates integration of a pretrained Large Language Model. ``Zero-shot'' denotes evaluation without task-specific fine-tuning. Dashes (-) indicate unreported metrics. The \textbf{best} and \underline{second-best} results are highlighted.}
\label{tab:reasonseg-tab}
\resizebox{0.65\linewidth}{!}{%
\begin{tabular}{lcccccccc}
\toprule
\multirow{2}{*}{Method} & \multirow{2}{*}{Use LLM} & \multirow{2}{*}{Zero-shot} & \multicolumn{3}{c}{ReasonSeg} & \multicolumn{3}{c}{LLM-Seg40k} \\
\cmidrule(lr){4-6} \cmidrule(lr){7-9}
& & & gIoU & cIoU & ncIoU & gIoU & cIoU \\
\midrule
GRES & $\times$ & $\checkmark$ & 22.4 & 19.9 & - & 14.2 & 15.9 \\
LISA-7B & $\checkmark$ & $\checkmark$ & 44.4 & 46.0 & - & 33.2 & 38.0 \\
LISA-13B & $\checkmark$ & $\checkmark$ & 48.9 & 46.9 & - & - & - \\
LLM-Seg-7B & $\checkmark$ & $\checkmark$ & 47.4 & 35.4 & 43.5 & 36.0 & 39.4 \\
\midrule
LISA-7B & $\checkmark$ & $\times$ & 52.9 & 54.0 & - & 37.6 & 48.5 \\
LISA-13B & $\checkmark$ & $\times$ & \underline{56.2} & \textbf{62.9} & \underline{52.8} & - & - \\
LLM-Seg-7B & $\checkmark$ & $\times$ & 55.4 & 45.6 & 51.0 & \underline{45.5} & \textbf{54.2} \\
GEAR-Seg (Ours) & $\checkmark$ & $\checkmark$ & \textbf{57.5} & \underline{47.5} & \textbf{54.0} & \textbf{52.2} & \underline{52.9} \\
\bottomrule
\end{tabular}
}
\end{table*}

\subsubsection{Generalization on Long-Tail Tasks.}
\label{subsubsec:referring}
To validate the Referring Segmentation Mode and inherent generalization in long-tail domains, we extend our zero-shot evaluation to complex agricultural scenarios.

\textbf{Instance Segmentation.} We evaluate zero-shot instance segmentation on three fruit datasets: StrawDI\_Db1 \cite{StrawDi}, Mega\_Blueberry \cite{sdmd2026}, and Mega\_Peach \cite{sdmd2026}, reporting COCO metrics against Grounded SAM \cite{groundedsam_2024}, YOLO-World \cite{yolo-world_2024}, and SDM \cite{sdmd2026}. 

\begin{figure}[t]
    \centering
    \includegraphics[width=1\linewidth]{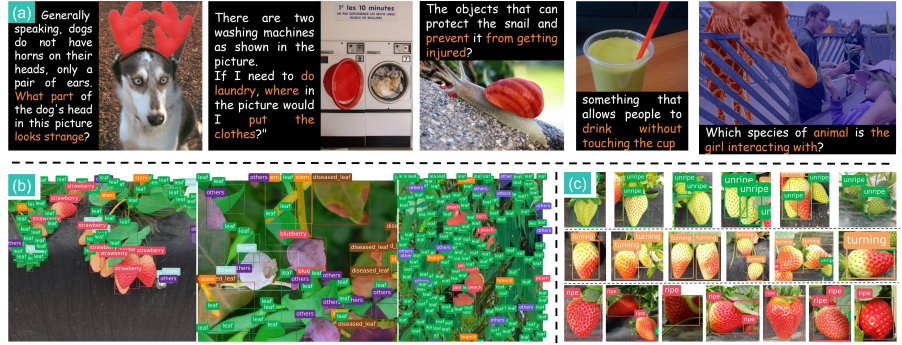}
    \caption{Qualitative results of the GEAR-Seg agent. (a) Complex reasoning segmentation on ReasonSeg and LLM-Seg40k. (b) Open-world auto-label extraction across diverse agricultural scenes, showcasing the zero-shot discovery of long-tail categories. (c) Fine-grained maturity grading, demonstrating precise attribute-based grounding under severe occlusion.}
    \label{fig:visual-results}
\end{figure}

As shown in \cref{tab:instance-seg}, GEAR-Seg demonstrates robust perception. In the stringent mAP$_{50:95}$ metric, it outperforms the second-best method by 13.2, 10.0, and 4.8 percentage points on StrawDI\_Db1, Mega\_Blueberry, and Mega\_Peach, respectively. Beyond standard prompting, GEAR-Seg autonomously discovers and assigns semantic labels to identifiable instances. As illustrated in \cref{fig:visual-results}(b), it successfully discovers long-tail categories like ``diseased leaf,'' underscoring its potential in unstructured environments.

\begin{table}[htbp]
\centering
\caption{Zero-shot fruit instance segmentation results. The \textbf{best} and \underline{second-best} results are highlighted.}
\label{tab:instance-seg}
\resizebox{0.85\linewidth}{!}{%
\begin{tabular}{lcccccccccc}
\toprule
\multirow{2}{*}{Method} & \multicolumn{3}{c}{StrawDI\_Db1} & \multicolumn{3}{c}{Mega\_Blueberry} & \multicolumn{3}{c}{Mega\_Peach} \\
\cmidrule(lr){2-4} \cmidrule(lr){5-7} \cmidrule(lr){8-10}
& mAP$_{50:95}$ & mAP$_{50}$ & mAR & mAP$_{50:95}$ & mAP$_{50}$ & mAR & mAP$_{50:95}$ & mAP$_{50}$ & mAR \\
\midrule
Grounded SAM & 0.242 & 0.329 & 0.425 & 0.297 & 0.328 & \textbf{0.706} & \underline{0.350} & 0.447 & \underline{0.617} \\
YOLO-World & 0.108 & 0.210 & 0.211 & 0.303 & \underline{0.380} & 0.542 & 0.163 & 0.230 & \textbf{0.630} \\
SDM & \underline{0.374} & \underline{0.429} & \underline{0.586} & \underline{0.320} & 0.348 & \underline{0.607} & 0.320 & \underline{0.460} & 0.577 \\
GEAR-Seg (Ours) & \textbf{0.506} & \textbf{0.594} & \textbf{0.642} & \textbf{0.420} & \textbf{0.476} & 0.595 & \textbf{0.368} & \textbf{0.473} & 0.585 \\
\bottomrule
\end{tabular}
}
\end{table}

\textbf{Fine-Grained Maturity Grading.} To test attribute-based reasoning, we introduce a fine-grained state classification task using 100 images (478 instances) from StrawDI. The model must classify targets into: unripe, turning, and ripe. This demands a synergy of language comprehension, physical perception, and robust dense segmentation. By seamlessly integrating LLM deduction with high-fidelity segmentation, GEAR-Seg overcomes the bottlenecks of existing open-vocabulary models, achieving a robust zero-shot accuracy of 83.2\%. As visually validated in \cref{fig:visual-results}(c), it performs reliably under severe occlusion.

\subsection{Ablation on Modularity and Scalability}
\label{subsec:ablation}

To systematically validate the structural advantages of our explicitly decoupled paradigm, we ablate the core text representation module and evaluate the framework's plug-and-play cognitive scalability.

\textbf{Necessity of Dense Descriptions.} A central premise of GEAR-Seg is that implicit reasoning requires fine-grained, attribute-rich representations. To prove this, we replace the DAM module with a general-purpose VLM (LLaMA) to generate mask descriptions. As shown in \cref{tab:ablation-modularity}, this substitution causes a drastic performance collapse on ReasonSeg (57.4 $\rightarrow$ 35.3 gIoU). Standard VLM captions inherently lack the granular physical and spatial details essential for resolving complex, manipulation-related queries, confirming that DAM's dense descriptions are the indispensable bedrock of our architecture.

\textbf{Plug-and-Play Cognitive Flexibility.} Unlike end-to-end black-box architectures that implicitly entangle perception and deduction, our decoupled nature allows for zero-cost cognitive upgrades without retraining. As detailed in \cref{tab:ablation-modularity}, effortlessly substituting the local Qwen3-32B reasoning core with the Gemini-3.1-Pro API directly boosts zero-shot performance to \textbf{58.2 gIoU} on ReasonSeg, establishing a highly flexible and competitive scaling path alongside concurrent SOTAs.

\textbf{Scalability for Edge Devices.} While heavy LLMs are indispensable for implicit commonsense reasoning, deploying them in resource-constrained environments is impractical. To validate cognitive on-demand capabilities, we replace the massive LLM with lightweight local models (Qwen-8B/30B~\cite{yang2025qwen3}) and a non-generative text encoder (Sentence-Transformers~\cite{reimers2019sentence}) for explicit agricultural referring tasks. As detailed in \cref{tab:ablation-llm}, thanks to the high-quality dense descriptions provided by DAM, lightweight models seamlessly integrate into the framework. Remarkably, Qwen-8B marginally outperforms DeepSeek-R1 on the StrawDI dataset (e.g., 0.512 vs. 0.507 in mAP$_{50:95}$). This implies that for direct attribute-matching tasks, smaller models effectively bypass the parsing complexities introduced by the overly verbose reasoning chains of massive LLMs. This dynamic scalability ensures optimal deployment for edge robotics.

\begin{table}[htbp]
\centering
\caption{Ablation on modularity and plug-and-play flexibility. We report zero-shot performance on ReasonSeg and LLM-Seg40k. The \textbf{best} results are highlighted.}
\label{tab:ablation-modularity}
\resizebox{0.65\linewidth}{!}{%
\begin{tabular}{lccccc}
\toprule
\multirow{2}{*}{Method Configuration} & \multicolumn{3}{c}{ReasonSeg} & \multicolumn{2}{c}{LLM-Seg40k} \\
\cmidrule(lr){2-4} \cmidrule(lr){5-6}
 & gIoU & cIoU & ncIoU & gIoU & cIoU \\
\midrule
SAM + LLaMA + DeepSeek-R1 & 35.3 & 17.7 & 28.1 & 30.4 & 26.2 \\
SAM + DAM + Qwen3-32B & 53.9 & 50.2 & 52.5 & 48.9 & 46.4 \\
SAM + DAM + DeepSeek-R1 & 57.4 & 47.5 & 54.0 & \textbf{52.2} & \textbf{52.9} \\
SAM + DAM + Gemini-3.1-Pro & \textbf{58.2} & \textbf{49.6} & \textbf{57.9} & 48.5 & 48.0 \\
\bottomrule
\end{tabular}
}
\end{table}

\begin{table}[htbp]
\vspace{-4mm} 
\centering
\caption{Scalability analysis of the cognitive module on agricultural referring tasks. DeepSeek-R1 is substituted with lightweight local models to evaluate plug-and-play robustness. The \textbf{best} and \underline{second-best} results are highlighted.}
\label{tab:ablation-llm}
\resizebox{0.85\linewidth}{!}{%
\begin{tabular}{lcccccccccc}
\toprule
\multirow{2}{*}{Cognitive Brain} & \multicolumn{3}{c}{StrawDI} & \multicolumn{3}{c}{Mega\_Blueberry} & \multicolumn{3}{c}{Mega\_Peach} \\
\cmidrule(lr){2-4} \cmidrule(lr){5-7} \cmidrule(lr){8-10}
& mAP$_{50:95}$ & mAP$_{50}$ & mAR & mAP$_{50:95}$ & mAP$_{50}$ & mAR & mAP$_{50:95}$ & mAP$_{50}$ & mAR \\
\midrule
DeepSeek-R1 (685B) & \underline{0.507} & \underline{0.595} & 0.641 & \textbf{0.420} & \textbf{0.476} & \textbf{0.595} & \textbf{0.368} & \textbf{0.472} & \textbf{0.585} \\
Qwen (8B, local) & \textbf{0.512} & \textbf{0.605} & \textbf{0.653} & \underline{0.395} & \underline{0.442} & \underline{0.543} & 0.345 & 0.439 & 0.564 \\
Qwen (30B, local) & 0.488 & 0.573 & \underline{0.650} & 0.379 & 0.424 & 0.523 & \underline{0.358} & \underline{0.457} & 0.580 \\
Sentence-TransF. & 0.456 & 0.546 & 0.622 & 0.391 & 0.410 & 0.507 & 0.353 & 0.435 & \underline{0.582} \\
\bottomrule
\end{tabular}
}
\end{table}

\subsection{Cascading Error and Efficiency Analysis}
\label{subsec:error_efficiency}

\begin{figure}[t]
    \centering
    \includegraphics[width=0.85\linewidth]{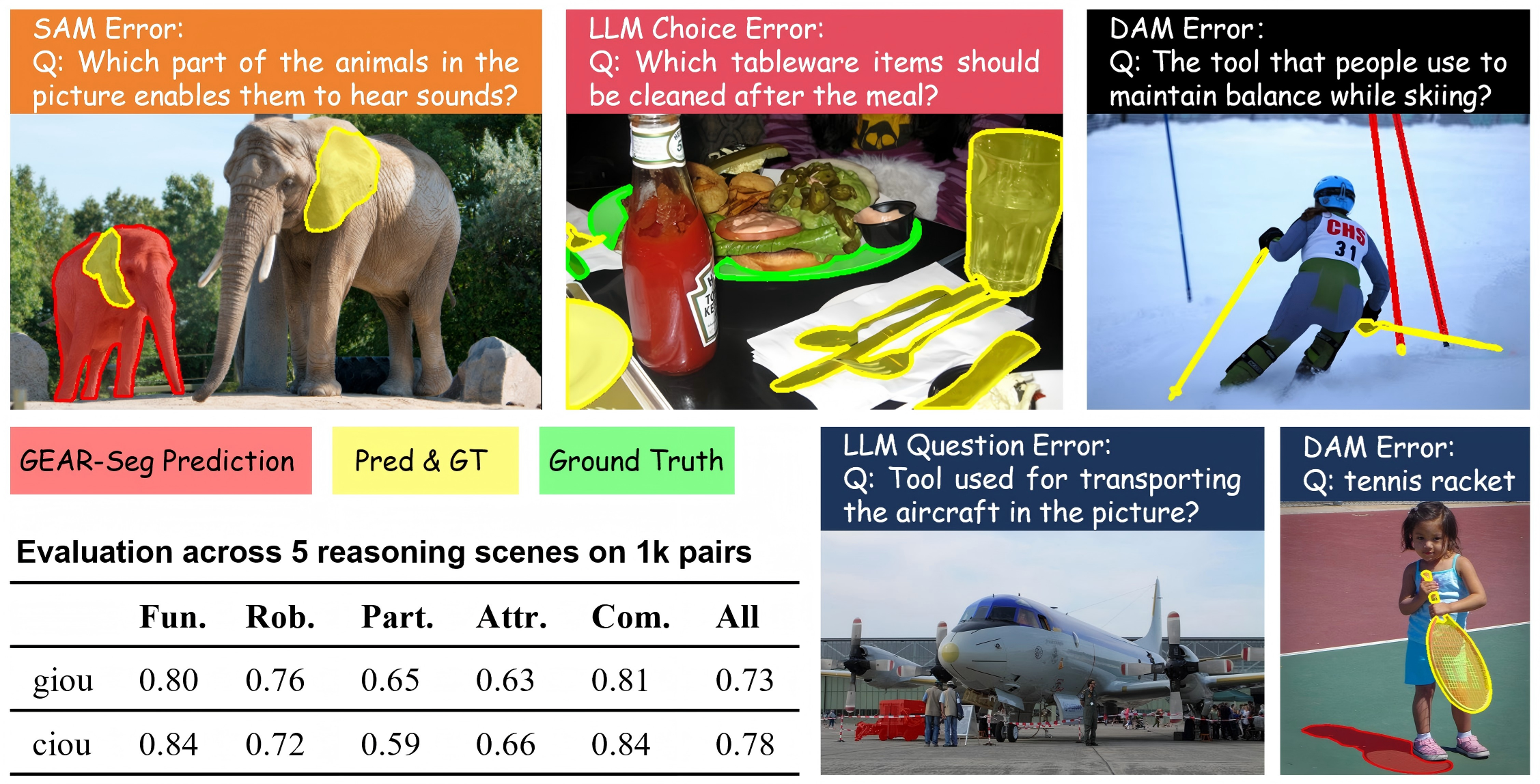}
    \caption{Accuracy evaluation and typical failure modes of the GEAR-Seg agent, illustrating cascading errors in complex reasoning and attribute hallucination.}
    \label{fig:failure_modes}
\end{figure}

\textbf{Cascading Error Analysis.} As a modular agent, GEAR-Seg inherently encounters cascading errors. To rigorously quantify dataset noise and algorithmic bottlenecks, we manually sampled and analyzed 2,000 QA-mask pairs (1,000 from the reasoning-heavy GEAR-131K and 1,000 from the long-tail StrawDI dataset). As detailed in \cref{tab:cascading-error}, SAM's class-agnostic perception remains highly stable across domains. However, downstream failure modes diverge significantly based on task complexity. In GEAR-131K, LLM deduction and choice errors dominate (combined $51.25\%$), reflecting the inherent difficulty of complex semantic reasoning. Conversely, in long-tail dense segmentation (StrawDI), DAM hallucination becomes the primary bottleneck ($78.93\%$). This explicitly underscores that mitigating Large Vision-Language Model (LVLM) hallucinations is critical for reliable fine-grained perception~\cite{zhu2025popen}. Typical failure modes are visualized in \cref{fig:failure_modes}.

\begin{table}[htbp]
\centering
\caption{Quantitative cascading error analysis on 2,000 manually sampled pairs. The breakdown isolates the source of failure across the decoupled modules.}
\label{tab:cascading-error}
\resizebox{0.7\linewidth}{!}{%
\begin{tabular}{lccccc}
\toprule
\multirow{2}{*}{Dataset} & \multirow{2}{*}{Accuracy} & \multicolumn{4}{c}{Error Breakdown} \\
\cmidrule(lr){3-6}
 & & SAM & DAM & LLM Question & LLM Choice \\
\midrule
GEAR-131K & 0.73 (gIoU) & 12.46\% & 36.28\% & 15.03\% & 36.22\% \\
StrawDI   & 0.66 (mAP$_{50}$) & 13.26\% & 78.93\% & / & 7.81\% \\
\bottomrule 
\end{tabular}
}
\end{table}

\textbf{Efficiency and Amortized Cost.} A common critique of agentic frameworks is inference latency. For GEAR-Seg, reasoning a highly dense 60-mask image on an NVIDIA RTX 4090 requires $\sim$34.3s (SAM: 2.3s/img, DAM: 0.5s/mask). While seemingly intensive for real-time edge deployment, this explicitly decoupled translation functions as a \textit{one-time} caching mechanism. Once the dense textual representation $\mathcal{D}$ is generated, the LLM can instantly resolve an unlimited number of diverse queries for the same image without any redundant visual re-computation. This renders the amortized computational cost highly efficient, making it a highly worthwhile tradeoff for offline dataset generation and complex multi-turn reasoning tasks.

\subsection{Knowledge Distillation to End-to-End Models}
To fully unleash the potential of the massive datasets generated by our data engine, we conduct knowledge distillation experiments. By transferring the dense cognitive capabilities of GEAR-Seg into end-to-end architectures, we aim to obtain lightweight models suitable for efficient real-time deployment.

\subsubsection{Distillation and SFT Utility for Reasoning Segmentation.}
We rigorously evaluate the distillation efficacy across two dimensions: the capability to substitute human annotation, and the utility of our generated logic chains in teaching semantic comprehension. We select LISA-13B~\cite{lisa2024} as the student architecture.

\textbf{Substituting Manual Annotation.} We first evaluate on the public ReasonSeg dataset to benchmark label quality. As reported in \cref{tab:dis1-reasonseg}, when trained on identical 239 images, the GEAR-Seg pseudo-label supervised model recovers a remarkable 92.4\% of the gIoU achieved by the strictly human-annotated baseline (51.4 vs. 55.6). Crucially, by effortlessly expanding the training data to 1.2k generated images, the distilled model substantially surpasses the human upper-bound across all metrics (e.g., 57.2 gIoU). This proves that high-fidelity automated supervision from our engine can effectively substitute costly manual annotation via scaled data volumes.

\textbf{SFT Utility of Logic Explanations.} To address whether our explicit reasoning chains actively teach semantic deduction, we ablate the textual loss ($\mathcal{L}_{txt}$) on the full 37.4k GEAR-131K training set. For a more rigorous evaluation, we expanded our manual verification to a highly challenging 1.8k test split. As detailed in \cref{tab:distill-sft}, training LISA-13B solely with mask supervision ($\mathcal{L}_{mask}$) yields a baseline gIoU of 26.6. However, incorporating our explicit explanation chains to co-optimize the native text objective ($\mathcal{L}_{txt}$) triggers a massive performance leap, reaching \textbf{37.2 gIoU}. This unequivocally demonstrates that our automated pipeline teaches genuine logical reasoning rather than merely fitting spatial pseudo-masks.

\begin{table}[htbp]
  \centering
  \setlength{\tabcolsep}{2.5pt} 
  \renewcommand{\arraystretch}{0.85}
  
  \begin{minipage}[t]{0.47\linewidth}
    \centering
    \caption{Distillation on ReasonSeg val set. Comparing GEAR-Seg pseudo-labels vs. Human annotations using LISA-13B.}
    \label{tab:dis1-reasonseg}
    \scriptsize 
    \begin{tabular}{llccc}
    \toprule
    \begin{tabular}[c]{@{}l@{}}Training \\ Images\end{tabular} 
    & \begin{tabular}[c]{@{}l@{}}Label \\ Source\end{tabular} 
    & gIoU & cIoU & ncIoU \\
    \midrule
    239 & Human & 55.6 & 57.0 & 53.4 \\
    239 & GEAR-Seg & 51.4 & 55.4 & 50.8 \\
    1.2k & GEAR-Seg & \textbf{57.2} & \textbf{63.6} & \textbf{55.4} \\
    \bottomrule
    \end{tabular}
  \end{minipage}\hfill
  \begin{minipage}[t]{0.47\linewidth}
    \centering
    \caption{SFT Utility on GEAR-131K test set. Explanations ($\mathcal{L}_{txt}$) significantly boost reasoning capabilities.}
    \label{tab:distill-sft}
    \scriptsize
    \begin{tabular}{llccc}
    \toprule
    \begin{tabular}[c]{@{}l@{}}Training \\ Data\end{tabular} 
    & \begin{tabular}[c]{@{}l@{}}Explanations\\($\mathcal{L}_{txt}$)\end{tabular} 
    & gIoU & cIoU & ncIoU \\
    \midrule
    37.4k & $\times$ (Mask Only) & 26.6 & 24.6 & 24.1 \\
    37.4k & $\checkmark$ (Full Loss) & \textbf{37.2} & \textbf{36.3} & \textbf{36.1} \\
    \bottomrule
    \end{tabular}
  \end{minipage}
\end{table}

\subsubsection{Distillation for Instance Segmentation.}
To validate practical utility for resource-constrained robotics, we deploy YOLOv8s for long-tail fruit instance segmentation. We compare lightweight student models supervised entirely by pseudo-labels from various automated pipelines against a rigorous human-annotated upper bound.

As detailed in \cref{tab:instance-seg-distill}, the GEAR-Seg supervised model substantially outperforms competing automated methods (e.g., Grounded SAM and YOLO-World). Remarkably, without any human intervention, our distilled model recovers up to 93.2\% of the rigorous manual baseline's performance (e.g., mAP$_{50:95}$ on Mega\_Blueberry). These results confirm GEAR-Seg's capability to supply highly accurate, scalable supervision for ultra-efficient edge models.

\begin{table}[htbp]
\centering
\caption{Zero-shot distillation results for fruit instance segmentation using YOLOv8s. The manual label baseline serves as the upper bound (*). The \textbf{best} and \underline{second-best} results among automated methods are highlighted.}
\label{tab:instance-seg-distill}
\resizebox{0.85\linewidth}{!}{%
\begin{tabular}{lcccccccccc}
\toprule
\multirow{2}{*}{Label Source} & \multicolumn{3}{c}{StrawDI\_Db1} & \multicolumn{3}{c}{Mega\_Blueberry} & \multicolumn{3}{c}{Mega\_Peach} \\
\cmidrule(lr){2-4} \cmidrule(lr){5-7} \cmidrule(lr){8-10}
& mAP$_{50:95}$ & mAP$_{50}$ & mAR & mAP$_{50:95}$ & mAP$_{50}$ & mAR & mAP$_{50:95}$ & mAP$_{50}$ & mAR \\
\midrule
Manual * & 0.773 & 0.935 & 0.790 & 0.766 & 0.897 & 0.815 & 0.680 & 0.912 & 0.741 \\
Grounded SAM & 0.443 & 0.567 & 0.600 & \underline{0.662} & 0.776 & 0.750 & 0.506 & 0.705 & 0.632 \\
YOLO-World & 0.278 & 0.396 & 0.377 & 0.623 & \underline{0.805} & \underline{0.754} & 0.545 & 0.761 & 0.665 \\
SDM & \underline{0.636} & \underline{0.774} & \textbf{0.699} & 0.596 & 0.699 & 0.730 & \underline{0.580} & \underline{0.792} & \underline{0.680} \\
GEAR-Seg (Ours) & \textbf{0.671} & \textbf{0.788} & \underline{0.685} & \textbf{0.714} & \textbf{0.873} & \textbf{0.855} & \textbf{0.611} & \textbf{0.809} & \textbf{0.688} \\
\bottomrule
\end{tabular}
}
\end{table}

\section{Conclusion}
We introduced GEAR-Seg, a grounded and explainable agent-based framework shifting reasoning segmentation from black-box entanglement to trackable, logic-driven deduction. GEAR-Seg delivers highly competitive zero-shot performance across five challenging benchmarks and serves as a scalable automated engine for GEAR-131K. This multifaceted dataset is specifically tailored for complex real-world interactions and diverse manipulation-related tasks. Furthermore, our distillation experiments demonstrate that GEAR-Seg effectively empowers lightweight, edge-deployable models with dense cognitive capabilities, successfully bridging the gap between sophisticated high-level reasoning and the rigorous practical constraints of resource-limited real-world applications.

\noindent\textbf{Limitations and Future Work.} As a modular agent, GEAR-Seg inevitably encounters cascading errors—a characteristic challenge inherent to the agentic paradigm—where initial perception inaccuracies regarding small or ambiguous targets can propagate through the subsequent reasoning chain. Addressing these fine-grained perception bottlenecks and further enhancing the robustness of pixel-to-text modality translation represent critical avenues for future research. Ultimately, we hope the GEAR-Seg framework and the GEAR-131K benchmark will serve as foundational catalysts for the next generation of interpretable vision systems in complex real-world applications and human-centric interactions.


\section*{Code and Data Availability}
The source code and the GEAR-131K dataset are publicly available at \url{https://github.com/AgRoboticsResearch/GEAR-Seg.git}.

\section*{Acknowledgements}
We sincerely thank the anonymous reviewers and Area Chairs for their constructive comments that helped improve this paper. This work was supported by the Fundamental Research Funds for the Central Universities (Grant No. 226-2025-00074). 

%
%
\bibliographystyle{splncs04}
\bibliography{main}

\clearpage        
\appendix

\section{Supplementary Material}

The supplementary material provides additional qualitative results and implementation details to support our proposed \textbf{GEAR-Seg}. 
Specifically, the supplementary archive is organized to present the following contents:

\begin{itemize}
    \item \textbf{Additional Dataset Visualizations:} Further qualitative examples demonstrating the linguistic diversity and reasoning complexity of the GEAR-131K dataset (Section~\ref{sec:dataset_vis}).
    \item \textbf{Zero-Shot Auto-Labeling Results:} Additional visualizations highlighting GEAR-Seg's open-ended auto-label extracting capability on long-tail agricultural datasets (Section~\ref{sec:auto-label}).
\end{itemize}

These materials complement the main text, providing deeper insights into both the data generation pipeline and the zero-shot generalization ability of our approach.

\subsection{Additional Dataset Visualizations}
\label{sec:dataset_vis}

As discussed in the main paper (Section 4), the proposed GEAR-131K dataset scales to a massive, instruction-tuning-ready resource of over 656K QA-mask pairs through a 5-fold linguistic expansion strategy. To provide a more concrete understanding of this data generation pipeline and the overall dataset quality, this section presents specific text examples and further qualitative visualizations.

\begin{figure*}[t]
\centering
\includegraphics[width=0.89\textwidth]{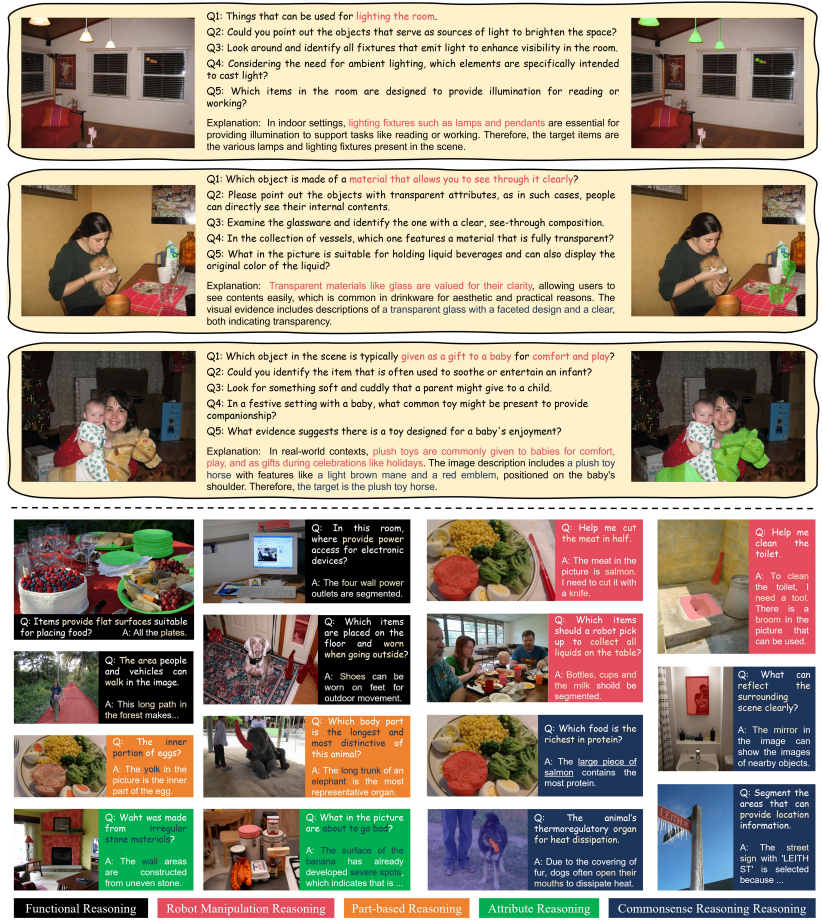}
\caption{
\textbf{Top:} Representative examples of the 5-fold linguistic expansion in the GEAR-131K dataset.
\textbf{Bottom:} Additional dataset visualizations of GEAR-131K.
}
\label{fig:linguistic-expansion}
\end{figure*}

First, the top panel of \cref{fig:linguistic-expansion} illustrates representative examples of our 5-fold linguistic expansion. For each reasoning target, we showcase how the underlying reasoning intent is rewritten into five semantically equivalent queries with diverse linguistic structures, accompanied by detailed explanations. This qualitative breakdown demonstrates how our strategy effectively enriches instruction diversity and reduces template bias. 

Second, the bottom panel of \cref{fig:linguistic-expansion} presents additional visual examples from the GEAR-131K dataset. These samples span a diverse set of real-world scenes and illustrate the variety of reasoning scenarios covered by the dataset. In particular, they highlight complex object interactions and contextual cues that require different levels of semantic reasoning to correctly ground the target objects. Such diversity further demonstrates the richness and practical relevance of GEAR-131K as a benchmark for reasoning-driven segmentation.

\subsection{Auto-Label Extracting Capability}
\label{sec:auto-label}

Beyond the referring segmentation tasks evaluated in the main text (Section 5.1), we also test GEAR-Seg's zero-shot generalization capabilities on long-tail data within complex agricultural environments. We design an open-ended auto-labeling experiment to demonstrate this. Specifically, given some unannotated images, we provide GEAR-Seg with a primary scene keyword (e.g., "strawberry" or "blueberry") and prompt the model to automatically discover and extract all other relevant semantic elements in the scene. Following this instruction, GEAR-Seg independently analyzes the global visual context, identifies fine-grained categories, and assigns appropriate text labels to all instances.

To ensure a rigorous and unbiased zero-shot evaluation, we conduct this experiment on several diverse sub-datasets from the MegaFruits Dataset (e.g., MegaBlueberry, MegaPeach, and MegaStrawberry). Importantly, it was introduced after the training data cut-offs of foundational models like SAM and DAM, strictly eliminating the possibility of data leakage.

\begin{figure*}[t]
\centering
\includegraphics[width=\textwidth]{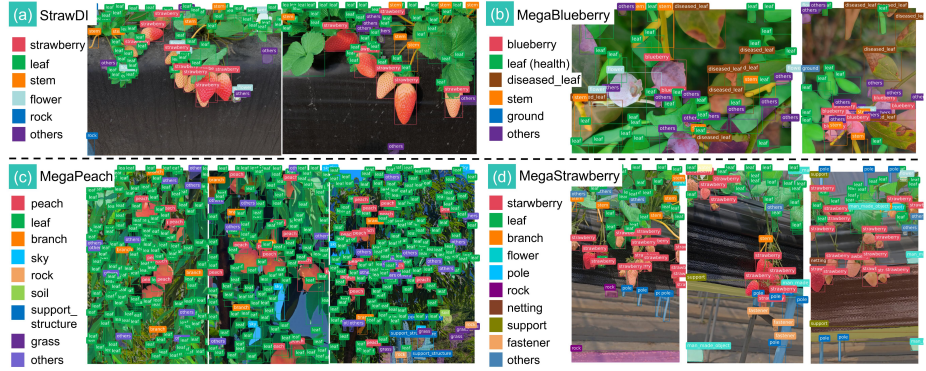}
\caption{
Examples of the auto-label extracting capability of GEAR-Seg across diverse MegaFruits datasets. By prompting the model to analyze the scene context, it automatically extracts a set of semantic categories and assigns fine-grained labels to each detected instance, including long-tail objects often missed by human annotators.
}
\label{fig:autolabel}
\end{figure*}

As illustrated in \cref{fig:autolabel}, GEAR-Seg successfully performs autonomous scene organization. It not only consistently grounds common agricultural elements (e.g., \textit{fruit}, \textit{leaf}, \textit{stem}), but critically, it discovers fine-grained, long-tail objects that are frequently overlooked during manual annotation. For instance, GEAR-Seg accurately identifies and labels \textit{diseased\_leaf} (\cref{fig:autolabel}b), as well as structural supporting elements like \textit{pole} and \textit{fastener} (\cref{fig:autolabel}d). This emergent ability to uncover subtle, unbounded semantic details demonstrates GEAR-Seg's significant potential for robust robotic manipulation and decision-making in unpredictable, real-world scenarios.

\end{document}